\documentclass[journal]{IEEEtran}
\usepackage{diagbox,multirow,epsfig,fbox,amsfonts,amsmath,multicol,enumitem,algorithm,pifont}
\usepackage{arydshln}
\usepackage{amssymb}
\usepackage{booktabs}
\usepackage[caption=false,font=footnotesize]{subfig}
\usepackage{array}

\usepackage{algpseudocode}
\usepackage[subfigure,titles]{tocloft}
\usepackage[numbers,sort&compress,comma]{natbib}

\graphicspath{{Figures/}}
\usepackage{hyperref}
\hypersetup{
	colorlinks=true,
	linkcolor=blue
}
\begin{document}
\title{GCD-DDPM: A Generative Change Detection Model Based on Difference-Feature Guided DDPM}
\author{Yihan Wen,
    Xianping Ma,
    Xiaokang Zhang,~\IEEEmembership{Member,~IEEE,}
   	and~Man-On Pun,~\IEEEmembership{Senior Member,~IEEE}
   	\thanks{This work was supported in part by the Guangdong Provincial Key Laboratory of Future Networks of Intelligence under Grant 2022B1212010001, and the National Natural Science Foundation of China under Grant 42371374 and 41801323. \textit{(Corresponding authors: Man-On Pun; Xiaokang Zhang)}}
    \thanks{Yihan Wen, Xianping Ma and Man-On Pun are with the School of Science and Engineering, The Chinese University of Hong Kong, Shenzhen, China (e-mail: wenyihan4396@gmail.com; xianpingma@link.cuhk.edu.cn; SimonPun@cuhk.edu.cn).}
    \thanks{Xiaokang Zhang is with the School of Information Science and Engineering, Wuhan University of Science and Technology, Wuhan 430081, China (e-mail: natezhangxk@gmail.com).}}
\maketitle

\begin{abstract}

Deep learning (DL)-based methods have recently shown great promise in bitemporal change detection (CD). Existing discriminative methods based on Convolutional Neural Networks (CNNs) and Transformers rely on discriminative representation learning for change recognition while struggling with exploring local and long-range contextual dependencies. As a result, it is still challenging to obtain fine-grained and robust CD maps in diverse ground scenes. To cope with this challenge, this work proposes a generative change detection model called GCD-DDPM to directly generate CD maps by exploiting the Denoising Diffusion Probabilistic Model (DDPM), instead of classifying each pixel into changed or unchanged categories. Furthermore, the Difference Conditional Encoder (DCE), is designed to guide the generation of CD maps by exploiting multi-level difference features. Leveraging the variational inference (VI) procedure, GCD-DDPM can adaptively re-calibrate the CD results through an iterative inference process, while accurately distinguishing subtle and irregular changes in diverse scenes. Finally, a Noise Suppression-based Semantic Enhancer (NSSE) is specifically designed to mitigate noise in the current step's change-aware feature representations from the CD Encoder. This refinement, serving as an attention map, can guide subsequent iterations while enhancing CD accuracy. Extensive experiments on four high-resolution CD datasets confirm the superior performance of the proposed GCD-DDPM. The code for this work will be available at \textit{https://github.com/udrs/GCD}.
\end{abstract}


\IEEEpeerreviewmaketitle

\section{Introduction}\label{sec:intro}

Change detection (CD) for multi-temporal remote sensing~(RS) imagery is a critical research area in urban and rural development investigation \cite{urban2}, natural hazard evaluation \cite{zhang2021semi}, and ecosystem observation \cite{CDforestRS}. Various CD techniques have been developed using image algebra, image transformation, and deep learning approaches. For instance, the image algebra-based approach extracts change magnitudes by exploiting either image differencing or change vector analysis (CVA) \cite{CVA3,CVA2019} whereas the image transformation-based approach aims to amplify the change information \cite{histogram69,lv2022object}. Recently, the deep learning (DL)-based approach has attracted much research attention for its outstanding performance in pixel classification. These DL-based CD methods often utilize deep neural networks to extract distinguishable features from the input bitemporal images by capitalizing on convolutional neural networks (CNNs) \cite{startConvolutional, startConvolution2, CNNstart11, CNNstart22}. In \cite{cnn2}, the Siamese CNN is employed to extract deep features, whereas \citet{FCC} introduced two Siamese extensions of fully connected CNNs to extract multi-scale features. Despite their good performance, these models are ineffective in preserving accurate detailed information due to the consecutive downsampling operations during the convolution process. To cope with this challenge, \citet{skipconnection} proposed a densely connected Siamese network to aggregate and refine features of multiple semantic levels, which helps to narrow semantic gaps and suppress localization errors. However, these CNN-based models are incapable of accurately extracting global interactions between contexts due to their limited receptive fields. Recently, the attention mechanisms have been introduced into CD to capture long-range dependencies and refine feature representations, thereby facilitating change map reconstruction \cite{attentionzhang, startattention1, startattention1223, zhangattnetion, LEVIRdataset}. For instance, \citet{attentionzhang} introduced a self-attention mechanism for CD by exploiting the Transformer architecture \cite{starttransformer, starttransformer2}, capturing abundant spatial-temporal relationships to obtain illumination-invariant features. In \cite{LEVIRdataset}, semantic segmentation and binary CD are integrated by utilizing a semantic-aware encoder and a temporal-symmetric transformer. However, it has been reported in \cite{CMT} that these models equipped with the self-attention mechanism are ineffective in capturing intricate local variations and edge details. As a result, these models cannot accurately characterize the change boundaries, which incurs prediction performance degradation. In summary, these methods utilize CNNs and Transformers as discriminative models to classify each pixel into changed or unchanged categories. 
However, since they rely on distinguishable feature representations for classification, they still struggle to simultaneously capture long-range dependencies while exploiting local spatial information.

In the meantime, the Denoising Diffusion Probabilistic Model (DDPM) has been recently proposed to enhance the generative capabilities of the diffusion model \cite{DDPM}. It has been shown that the DDPM outperforms the conventional Generative Adversarial Networks (GANs) and the Transformer in various applications, including super-resolution \cite{srdiff,srdiff3}, segmentation \cite{diffseg,diffseg2, segmentationDDPM}, inpainting \cite{impaint,impaint2,impaint3}, and conditional image generation \cite{imgenera,imgerate2}.  Specifically, a diffusion model can be trained through the variational inference (VI) procedure on an extensive collection of off-the-shelf remote sensing images to generate images that increasingly resemble authentic images over a finite time period. The resulting diffusion model progressively improves its semantic extraction capability, which is instrumental in acquiring accurate CD maps. Recently, a DDPM-based model called DDPM-CD \cite{DDPM-CD} has been developed for semantic segmentation and CD by employing the pre-trained DDPM as a feature extractor for multitemporal feature representation. However, upon integrating the DDPM-based feature extractor into the decoder and classifier, the DDPM-CD is essentially a discriminative model, following a single forward propagation approach. This setup overlooks the gradual refinement and generative capabilities inherent in the DDPM.

In this paper, we propose a pure generative CD model named GCD-DDPM. Unlike discriminative models that aim to classify each pixel into changed or unchanged categories, the proposed GCD-DDPM employs an end-to-end architecture to generate a CD map directly. Specifically, a Difference Conditional Encoder (DCE) is designed to extract multi-level difference information from pre- and post-change images, which is then integrated into the model's sampling process to guide the generation of CD maps. By leveraging the VI mechanism of diffusion models, i.e., introducing varying degrees of noise into the CD map and then methodically recovering the map, GCD-DDPM can learn distinguishable change-aware feature representations, enhancing its sensitivity to subtle data distinctions. Furthermore, a Noise Suppression-based Semantic Enhancer (NSSE) is explicitly designed to mitigate noise in the change-aware features and guide subsequent iterations. Unlike discriminative models that employ a single forward-propagation process, the GCD-DDPM implements an adaptive calibration approach, by continuously refining and improving the predictions of CD maps through an iterative inference process. The main contributions of this paper are as follows:

\begin{enumerate}
    \item The proposed GCD-DDPM is a pure generative model tailored for CD tasks. By utilizing an adaptive calibration approach, the GCD-DDPM excels in gradually learning and refining data representations, effectively distinguishing diverse changes in natural scenes and urban landscapes;
    \item The GCD-DDPM introduces differences among multi-level features extracted from pre- and post-change images within DCE, which are then integrated into the sampling process to guide the generation of CD maps. This method allows for a more fine-grained capture of changes;
    \item An encoder called NSSE is proposed by employing an attention mechanism with learnable frequency-domain filters to suppress noise in the difference information derived in the current step. This process is vital for aiding the DCE in extracting more accurate change-aware representations and subsequently, enhances the DCE's ability to distinguish and capture changes.
\end{enumerate}

The rest of this paper is organized as follows. Section~\ref{sec:related} describes the related work of deep learning-based CD methods and the recent DDPM-based models reported in the RS literature. After that, Section~\ref{sec:method} provides the details on the proposed GCD-DDPM, whereas extensive experimental results are presented in Section~\ref{sec:experiment}. Finally, the conclusion is given in Section~\ref{sec:conclusion}.

\section{Related work}\label{sec:related}
\subsection{CNN-based models}
Conventional CNN-based models have been widely used for extracting difference maps and features, primarily focusing on capturing local spatial information \cite{cnn,zhang2023federated,zhang2019land}. For example, a symmetric convolutional coupling network called SCCN \cite{SCCN} employed unsupervised learning to optimize a coupling function based on heterogeneous images, aiming to capture the intrinsic relationship with emphasis on the difference between input images. In addition, skip-connections were introduced in \cite{ReCNN} to handle temporal connections in multi-temporal data while extracting rich spectral-spatial features. However, these CNN-based methods have inherent limitations in modeling global dependencies and capturing comprehensive spatial-temporal information as they primarily focus on local regions. As a result, these CNN-based models exhibit poor performance in capturing complex patterns and changes that occur over larger areas of intricate spatial distributions.

\subsection{Transformer-based models}
In contrast, the Transformer architecture has demonstrated exceptional performance in various computer vision tasks, including CD, by effectively addressing the challenge of capturing long-range dependencies \cite{BIT,changeformer}. The use of self-attention modules in the Transformer architecture enables the capturing of global contextual relationships, which helps overcome the limited receptive field problem plaguing the CNN-based models. Several studies have built upon this success by introducing Transformer-based models tailored for CD. For instance, \citet{swinsunet} introduces a pure transformer network for CD, featuring a Siamese U-shaped structure with Swin Transformer blocks to extract multiscale features and capture global spatiotemporal information effectively. In \cite{wang2022spectral}, a transformative spectral transformer encoder is introduced to combine spectral, spatial, and temporal elements using a spatial transformer for spatial texture analysis. Meanwhile, \citet{AMTCD} merges ConvNets with transformers within a Siamese network, handling bi-temporal image contextual information. Moreover, \citet{li2022transunetcd} proposed TransUNetCD, which uniquely combines transformers and UNet for CD, adeptly capturing global contexts and deep features.

Despite their many advantages, the Transformer-based models suffer from inaccurate prediction of fine-grained details of CD maps due to the inherent characteristics of the Transformer architecture. As a result, the predicted CD maps exhibit relatively low precision and coarse estimation of edge details. Furthermore, while the Transformers excel at capturing global contextual information, they are less effective in capturing intricate local variations and edge details \cite{CMT}. This limitation hinders them from accurately delineating the boundaries of change regions.

\begin{figure}
\centering
\includegraphics[width=0.95\linewidth]{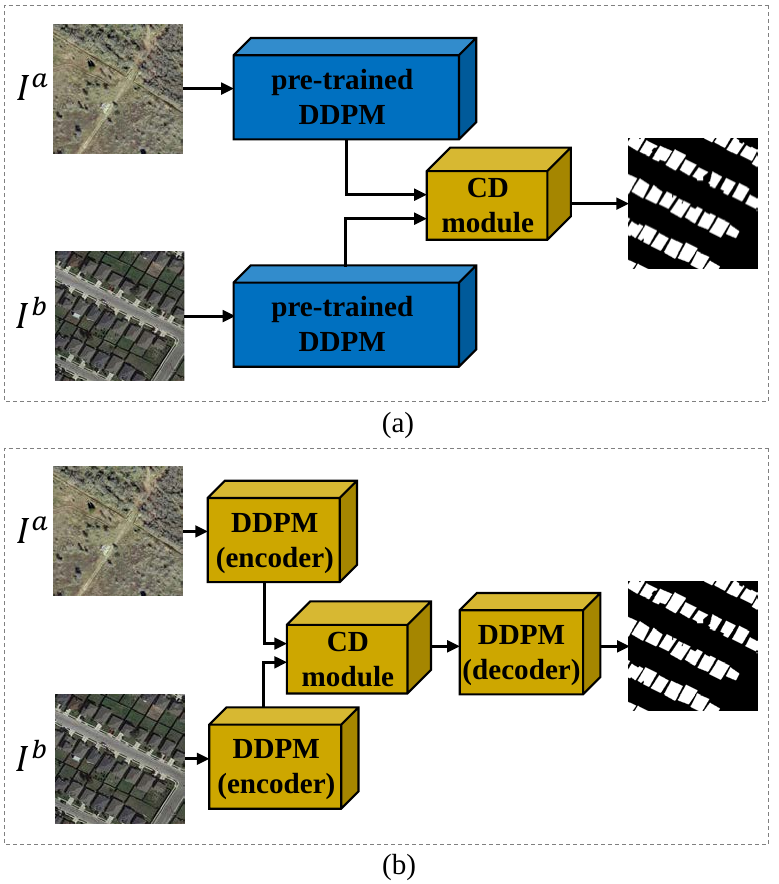}
\caption{A general comparison between (a) DDPM-CD and (b) the proposed GCD-DDPM. In DDPM-CD, DDPM is trained by remote sensing images in advance, and only the CD module is optimized in the CD task, while GCD-DDPM takes full advantage of the characteristics of the DDPM to train the entire model end-to-end. In the figure, the parameters of the blue blocks are fixed, while the orange blocks are optimized during the CD task.}
\label{fig:DDPMCD}
\end{figure}

\begin{figure*}
\centering
\includegraphics[width=1\linewidth]{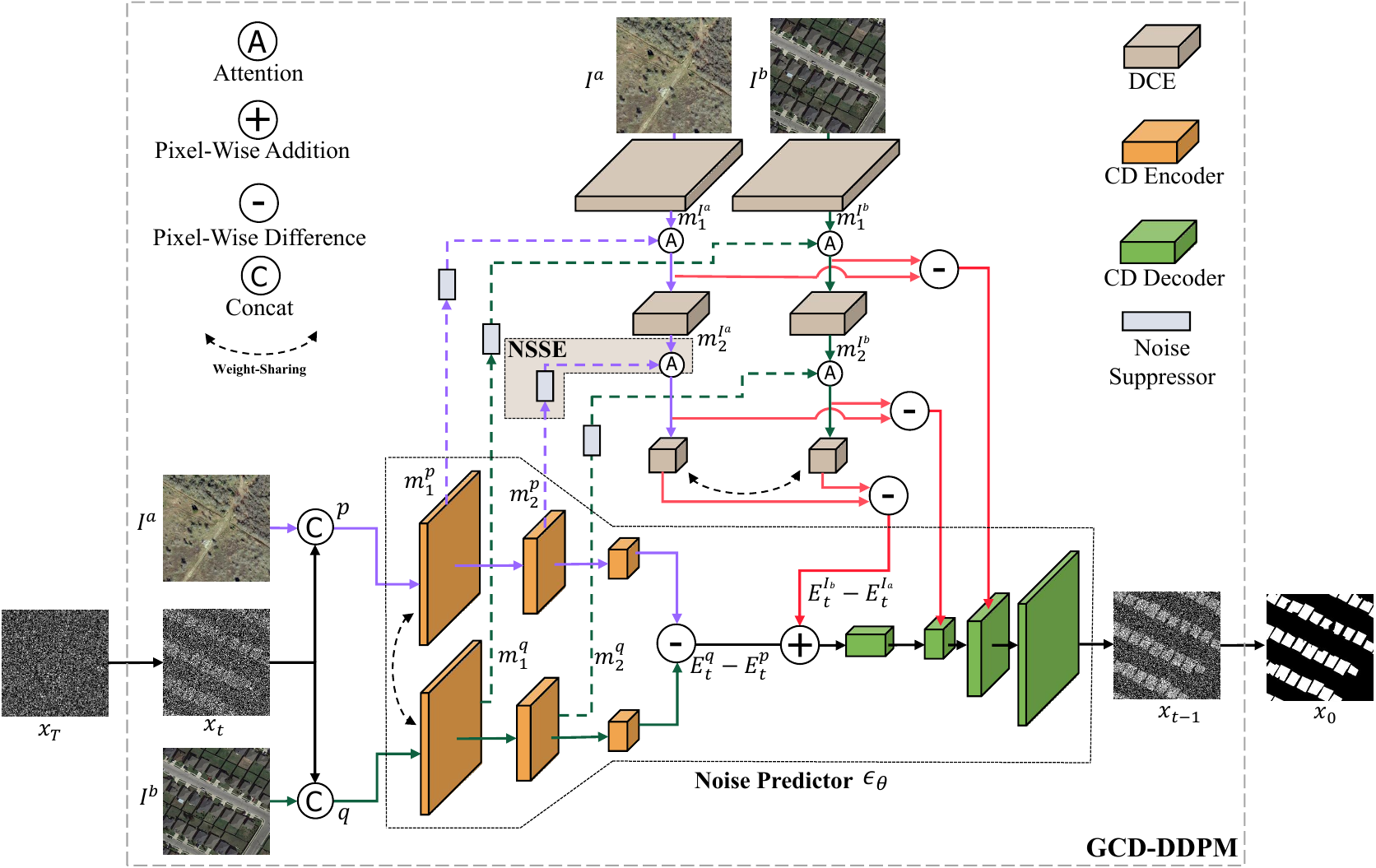}
\caption{An illustration of one timestep in the proposed GCD-DDPM, featuring Noise Predictor, NSSE modules and DCE. For clarity, only one NSSE module is presented in a gray box.}
\label{fig:modelstructure}
\end{figure*}

\subsection{DDPM-based models}
Diffusion models have achieved remarkable success in generating high-fidelity images and have recently been introduced into segmentation tasks \cite{diffmed}.
Compared to CNNs and Transformers, the DDPM-based generative models offer several advantages that enable them to capture complex data distributions, leading to accurate prediction of fine-grained details and edge information in CD maps. More specifically, the diffusion model employed in DDPM captures the diversity and complexity of input images by leveraging probabilistic modeling. Furthermore, the model can adapt to various scenes and image features, including the shape, size, and texture of ground objects, through learning to transform a standard normal distribution into an empirical data distribution. In remote sensing, \citet{chen2023spectraldiff} introduced SpectralDiff for hyperspectral image classification to capture spectral-spatial relationships and obtain context-rich features. Furthermore, the DDPM-CD model \cite{DDPM-CD} utilized the pre-trained DDPM as a feature extractor into which the remote sensing images are directly fed for feature extraction. These extracted segmentation features are subsequently channeled into a lightweight module for the CD task, which will be optimized while the DDPM is fixed.

However, the existing methods in CD tasks do not fully exploit the benefits of DDPM. They only utilize the pre-training of the diffusion model for unsupervised feature extraction and overlook its generative capability for change perception by leveraging its random sampling process. Therefore, fine-grained details of land cover changes cannot be well retained in CD maps. To address this drawback, we introduce GCD-DDPM endowed with VI to facilitate a more nuanced and robust probabilistic framework as compared to DDPM-CD. Furthermore, it employs an end-to-end training architecture that directly generates accurate CD maps through iterative refinements during the inference process. This transformative model adaptively captures the intrinsic complexities and variances present in CD scenarios, thereby substantially outperforming existing solutions in terms of both accuracy and adaptability. The differences between them are presented in Fig.~\ref{fig:DDPMCD} where we denote by $I^a\in\mathbb{R}^{H\times W \times 3}$ and $I^b\in\mathbb{R}^{H\times W \times 3}$ the pre- and post-change images, respectively.

\section{METHODOLOGY}\label{sec:method}
\subsection{Diffusion Process of GCD-DDPM}
The proposed GCD-DDPM is a generative model consisting of two stages, namely, the forward diffusion stage and the reverse diffusion stage. In the forward process, the CD label $x_0$ is gradually added with Gaussian noise through a series of steps $T$. During the reverse diffusion stage, a neural network is trained as a noise predictor to reverse the noising process and subsequently, recover the original data. 

\subsubsection{Forward Process}\label{sec:intro}
The diffusion process involves generating a series of data points $x_1$, $x_2$, ..., $x_T$ conditioned on a given initial data distribution $x_{0} \sim q\left(x_{0}\right)$. This process can be mathematically formulated as follows:
\begin{equation}
q(x_1, \ldots, x_T \mid x_0) = 
\prod_{t=1}^{T} q(x_t \mid x_{t-1}),
\end{equation}
where
\begin{equation}
q(x_t\mid x_{t-1}) = \mathcal{N}(x_t; \sqrt{1-\beta_{t}} x_{t-1}, \beta_t \mathbf{I}),    
\end{equation}
with the variance schedule ${ \beta_{1}, ..., \beta_{T} } \in \left(0, 1\right)$ consisting of a set of hyperparameters. Furthermore, $\mathcal{N}\left(\cdot;\cdot,\cdot\right)$ represents a Gaussian distribution, which plays a pivotal role in the diffusion process, gradually introducing noise into the data. The covariance matrix \( \beta_t \mathbf{I} \), where \( \mathbf{I} \) is the identity matrix, is designed to ensure that noise is added independently to each dimension of \( x_{t-1} \) with a variance of \( \beta_t \).

This recursive formulation represents a Gaussian distribution characterized by a mean of $\sqrt{1-\beta_{t}} x_{t-1}$ and a variance of $\beta_t \mathbf{I}$. Furthermore,the mathematical relationship between $x_{0}$ and $x_{t}$ is formulated as:
\begin{equation}
x_{t} = \sqrt{\overline{a}_{t}}x_{0} + \sqrt{1-\overline{a}_t}\epsilon,
\end{equation}
where 
\begin{equation}
\overline{a}_t = \prod_{s=1}^{t} \left(1 - \beta_s\right)
\end{equation}
with $\epsilon$ being modeled with the standard normal distribution of zero mean and unit variance. 

\subsubsection{Reverse Process}\label{sec:intro}
The reverse process involves transforming the latent variable distribution $p_{\theta}\left(x_{T}\right)$ into the data distribution $p_{\theta}\left(x_{0}\right)$ parameterized by $\theta$. This transformation is defined by a Markov chain featuring learned Gaussian transitions with the initial distribution modeled as the standard normal distribution $\mathcal{N}(x_T; 0, \mathbf{I})$. Mathematically, the transformation can be represented as:
\begin{equation}\label{eq:trans}
p_\theta(x_0, \dots, x_{T-1} | x_T) = \prod_{t=1}^T p_\theta(x_{t-1} | x_t),
\end{equation}
where
\begin{equation}\label{eq:cond}
p_\theta(x_{t-1} | x_t) = \mathcal{N}(x_{t-1}; \mu_\theta(x_t, t), \sigma_\theta^2(x_t, t) \mathbf{I}),
\end{equation}
with $\theta$ denoting the parameters governing this reverse process.

In the training phase, based on VI, the objective is to optimize these parameters $\theta$ such that the reverse diffusion process accurately approximates the original data distribution. To achieve this goal, we introduce a neural network-based noise predictor, denoted as \({NP}(\cdot; \theta)\). This function estimates the noise \( \epsilon \) for given \( x_t \) as follows:
\begin{equation}
\epsilon_{\theta} = {NP}(x_t; \theta).
\end{equation}
We adopt a mean squared error loss function \( \mathcal{L}(\theta) \), aiming to minimize the discrepancy between \( \epsilon \) and \( \epsilon_\theta \):
\begin{equation}
\mathcal{L}(\theta) = \nabla_{\theta} \left\| \epsilon - \epsilon_{\theta} \right\|^2.
\end{equation}

During the inference stage, we begin by sampling an initial \( x_T \) from the standard Gaussian distribution. Using \( x_T \) as the starting point, subsequent data points \( x_{T-1}, x_{T-2}, \ldots, x_0 \) are sampled recursively based on the learned transition models defined by Eq.~\eqref{eq:trans} and Eq.~\eqref{eq:cond} with
\begin{eqnarray}\label{eq:inference}
\mu_\theta(x_t, t) &=& \frac{1}{\sqrt{\alpha_t}}\left( x_t - \frac{\beta_t}{\sqrt{1 - \tilde{\alpha}_t}}  \epsilon_\theta(x_t, t)\right),\\
\sigma_\theta(x_t, t) &=& \sqrt{\tilde{\beta}_t},
\end{eqnarray}
for $t \in \left\{ T, T-1, \ldots, 1 \right\} $. The process is iteratively applied to reconstruct the noise image, ultimately yielding a clear segmentation during the inference phase.

\subsection{GCD-DDPM Network}\label{sec:intro}
Most existing DDPM-based methods utilized the DDPM as a feature extractor to extract multi-level features without fully exploiting the current estimate as prior knowledge during the inference process \cite{DDPM-CD}. In this work, we propose to generate high-quality CD maps directly through end-to-end training without the necessity of additional training on the diffusion models. The proposed network consists of two key components, namely a noise predictor denoted as $\epsilon_\theta$ built on a classical Encoder-Decoder architecture and a novel CD module, namely Difference Conditional Encoder (DCE). The CD Encoder in noise predictor is designed to effectively estimate noise features before the DCE seamlessly fuses the difference information derived from pre- and post-change images. In addition, to effectively leverage the multi-step iterative prediction capability of the diffusion model, we further propose an NSSE to integrate the current-step CD noise feature from the CD Encoder into the conditional encoding for enhancement. After that, multi-scale feature maps that contain change-aware information from the DCE are fused with the current-step CD noise feature through the pixel-wise addition and skip connections in the CD Decoder to obtain current-step CD-related noise, which will be leveraged as prior information for next-step iteration. By iteratively sampling the Gaussian noise, the proposed GCD-DDPM progressively improves the accuracy of the resulting CD map by accurately characterizing the difference between the input images. An illustration of the proposed GCD-DDPM is presented in Fig.~\ref{fig:modelstructure}. In the following sections, details about the Noise Predictor, DCE, and NSSE will be discussed.

\subsubsection{Noise Predictor}
In accordance with the standard implementation of the DDPM, a U-Net including CD Encoder and CD Decoder is employed as a noise predictor as shown in Fig.~\ref{fig:modelstructure}. Furthermore, the step noise estimation function is conditioned on raw images and takes the following form:
\begin{equation}
\epsilon_\theta(x_t, I^a, I^b, t) = \mathcal{D}\left((E^{I^b}_t-E^{I^a}_t)+(E^{q}_t-E^{p}_t),t\right),
\end{equation}
where $\mathcal{D}\left(\cdot\right)$ is the CD Decoder for final CD maps. Furthermore, $E^{I^b}_t$ and $E^{I^a}_t$ denote the conditional embedding features from DCE at step $t$ whereas $E^{q}_t$ and $E^{p}_t$ represent the noise features from the CD Encoder at step $t$. Thereby, $E^{I^b}_t-E^{I^a}_t$ is the difference of the bitemporal conditional embedding features while $E^{q}_t-E^{p}_t$ stands for the difference of the bitemporal noise features. Finally, $p$ and $q$ are the initial noise features in the noise predictor defined as follows:
\begin{eqnarray}
	p&\triangleq&Concat\left(x_{t}, I^{a}\right),\\
	q&\triangleq&Concat\left(x_{t}, I^{b}\right),
\end{eqnarray}
with $Concat$$\left(\cdot\right)$ being the concatenation operator. 

In the noise predictor's CD Encoder of the GCD-DDPM, the input images are processed through a flexible series of residual blocks (ResBlocks) with downsampling operations. The CD Decoder then upscales the features back to their original spatial dimensions through ResBlocks and upsampling layers and integrates them with the DCE module's outputs through skip connections. During this process, features from the corresponding levels of the DCE module are merged to enhance the detail and quality of the features. The final output of the CD Decoder is a detailed single-channel CD-related noise $\epsilon_{\theta}$. In accordance with Eq.~\eqref{eq:inference}, this noise is then utilized to generate the subsequent CD map.

\subsubsection{DCE}\label{sec:intro}
\begin{figure*}
\centering
\includegraphics[width=0.95\linewidth]{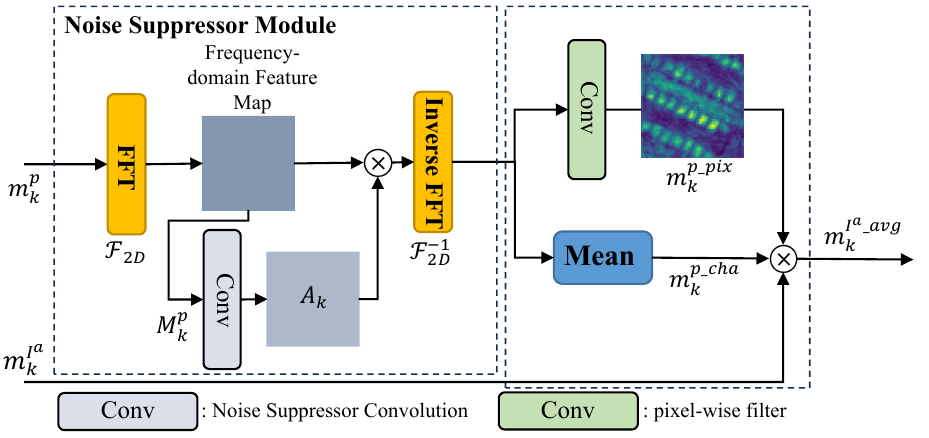}
\caption{An illustration of the proposed attentive module NSSE. The spatial image information is first transformed into the frequency domain using the Fast Fourier Transform (FFT). After that, an attentive-like mechanism is utilized to suppress high-frequency noise before the extracted CD information is converted back into the spatial domain using the Inverse Fast Fourier Transform (IFFT).}
\label{fig:modelstructure2}
\end{figure*}

The DCE module is developed to extract change information from each sample within the diffusion model framework. We denote by $m_{k}^{I^a}$ and $m_{k}^{I^b}$ the conditional feature maps derived from pre- and post-change images obtained in the $k$-th block, respectively. Despite that $m_{k}^{I^a}$ and $m_{k}^{I^b}$ contain raw target information without distortion, it is challenging to directly compute the difference from $\left\{m_{k}^{I^a},m_{k}^{I^b}\right\}$. In contrast, the current-step CD map, despite being less accurate, contains enhanced target regions. Inspired by this observation, this work proposes to integrate the current-step CD map information $x_t$ into DCE by extracting information from multi-level noise features. More specifically, the extracted multi-level conditional features $m_{k}^{I^a}$ and $m_{k}^{I^b}$ in DCE are integrated with the current-step corresponding noise features from the Nosie Predictor. Thus motivated, we propose the NSSE module to enhance and calibrate the conditional embedding features.

\subsubsection{NSSE}\label{sec:intro}
We show the $k$-th level data flow for pre-change features in NSSE, which are the same as post-change features. The proposed NSSE consists of a Noise-Suppressor module as shown in Fig.~\ref{fig:modelstructure2} designed to suppress noise inherent in $x_{t}$ by eliminating high-frequency noise using a parameterized attention map. The $k$-th level noise feature $m_{k}^{p}$ can be generated by passing initial noise features through the corresponding convolutional block within the CD Encoder architecture. After that, we transform $m_{k}^{p}$ into frequency-domain noise feature $M_{k}^{p}$ using a 2D Fast Fourier transform (FFT) along the spatial dimensions as follows:
\begin{equation}
    M_{k}^{p}= \mathcal{F}_{2D}(m_{k}^{p}),\label{eq17}
\end{equation}
where $\mathcal{F}_{2D}(\cdot)$ represents the 2D FFT function. 

Next, we multiply $M_{k}^{p}$ with a parameterized attentive map $A_{k}$ before converting the results back to the spatial domain using the inverse  2D FFT. Mathematically, the above process can be modeled as:
\begin{equation}
    \tilde{m}_{k}^{p} = \mathcal{F}_{2D}^{-1}(A_{k} \otimes M_{k}^{p}),\label{eq18}
\end{equation}
where $\mathcal{F}_{2D}^{-1}\left(\cdot\right)$ stands for the inverse FFT function while $\otimes$ stands for the element-wise product operator.

The noise suppression module discussed above can be considered as an adaptive frequency filter commonly employed in digital image processing. In sharp contrast to conventional spatial attention, the proposed module can learn to constrain the high-frequency component for adaptive integration by globally adjusting filtering frequencies.

After obtaining $\tilde{m}_{k}^{p}$, pixel-wise and channel-wise attention mechanisms are employed to facilitate the fusion process between $\tilde{m}_{k}^{p}$ and the conditional embedding $m_{k}^{I^a}$. Specifically, CD feature $\tilde{m}_{k}^{p}$ extracted from noise feature $p$ is fed into two separate convolution layers : 
\begin{eqnarray}
	m^{p\_pix}_{k}&=& Conv\left({\cal N}(\tilde{m}_{k}^{p})\right),\\\label{eq19}
	m^{p\_cha}_{k}&=& Mean\left({\cal N}(\tilde{m}_{k}^{p})\right),\label{eq20}
\end{eqnarray}
where the $Conv(\cdot)$ function here maps $\tilde{m}_{k}^{p}$ to a feature map $m^{p\_pix}_{k}$ with only one single channel, serving as a trainable pixel-wise filter. In addition, the $Mean(\cdot)$ function represents the global average pooling operation, yielding a one-dimensional vector $m^{p\_cha}_{k}$ of channel length. Furthermore, ${\cal N}(\cdot)$ stands for Linear Normalization. Finally, we can compute the augmented conditional embedding $m^{I_{a}\_aug}_k$ as follows:
\begin{eqnarray}
	m^{I_{a}\_aug}_k &=&  m^{p\_pix}_{k} \otimes m^{p\_cha}_{k} \otimes (m^{I^a}_k).\label{eq21}
\end{eqnarray}
For the $k$-th level post-change features, we can compute $m^{I_{b}\_aug}_k$ by the similar procedures from Eqs~\eqref{eq17}-\eqref{eq21}.

Finally, the multi-scale feature maps encapsulating differential information from the conditional encoders are seamlessly fused with the current-step CD features, denoted as $m^{I_{a}\_aug}_k - m^{I_{b}\_aug}_k$. This fusion is executed within the CD Decoder through pixel-wise addition operations and skip connections, as illustrated in Fig.~\ref{fig:modelstructure}. These fused features are then passed onto the last encoding layer before a decoder equipped with a residual network decodes the high-dimensional features. By iteratively sampling Gaussian noise, the proposed GCD-DDPM model incrementally enhances the precision of the resulting CD map by accurately characterizing the difference between the input images. The whole training and inference process of the GCD-DDPM are displayed in Algorithms~\ref{alg1} and~\ref{alg2}.

\begin{algorithm}
\caption{GCD-DDPM Training}
\begin{algorithmic}[1]
\Require Total diffusion steps $T$, image pairs $(I^a, I^b)$ and CD map $x_0$ within dataset $D$
\State Initialize $\theta$ with random weights
\For{each epoch}
        \State Sample $\{I^a_k, I^b_k, x_{0}\} \sim D$
        \State Sample $\epsilon \sim \mathcal{N}(0, \mathbf{I})$
        
        \State $\beta_t = \text{linear schedule from } 10^{-4} \text{ to } 2 \times 10^{-2}$
        \State $\alpha_t = 1 - \beta_t$
        \State $\bar{\alpha}_t = \prod_{s=0}^t \alpha_s$
        \State $x_t = \sqrt{\bar{\alpha}_t}{x_0} + \sqrt{1 - \bar{\alpha}_t} \epsilon$
        \State $\epsilon_{\theta}(x_t; \theta) = \epsilon_{\theta}(\sqrt{\bar{\alpha}_t}{x_0} + \sqrt{1 - \bar{\alpha}_t} \epsilon; \theta)$
        \State Take gradient descent $\nabla_{\theta} \| \epsilon - \epsilon_{\theta}(x_t; \theta) \|^2$
\EndFor
\end{algorithmic}
\label{alg1}
\end{algorithm}

\begin{algorithm}
\caption{GCD-DDPM Inference}
\begin{algorithmic}[1]
\Require Total diffusion steps $T$, trained parameters $\theta$, initial noise level
\State Sample $x_T \sim \mathcal{N}(0, \mathbf{I})$
\For{$t = T, T-1, \ldots, 1$}
    \State  $z \sim \mathcal{N}(0, \mathbf{I})$
    \State Linear variance schedule $\beta_t = \frac{10^{-4}(T - t) + 2 \times 10^{-2}(t - 1)}{T - 1}$ 
    \State $\alpha_t = 1 - \beta_t$
    \State $\bar{\alpha}_t = \prod_{s=0}^t \alpha_s$
    \State $\tilde{\beta}_t = \frac{1 - \overline{\alpha}_{t-1}}{1 - \overline{\alpha}_{t}} \beta_t$
    \State $\mu_\theta(x_t, t) = \frac{1}{\sqrt{\alpha_t}}\left( x_t - \frac{\beta_t}{\sqrt{1 - \tilde{\alpha}_t}}  \epsilon_\theta(x_t, I^a, I^b, t)\right)$
    \State $\sigma_\theta(x_t, t) = \sqrt{\tilde{\beta}_t}$
    \State Sample $x_{t-1} = \mu_\theta(x_t, t) + \sigma_\theta(x_t, t) z$
\EndFor
\State \Return $x_0$ as the denoised CD map
\end{algorithmic}
\label{alg2}
\end{algorithm}

\begin{table*}[!htb]
	\centering
	\caption{The structure of Noise Predictor. The parameters contain kernel size, stride and channel dimension}
	\label{table_network}
	\begin{tabular}{cccccc} 
		\toprule
		\multicolumn{3}{c}{{CD Encoder}}     & \multicolumn{3}{c}{{CD Decoder} }       \\
		{Stage}        & {Parameters} & {Output}  & {Stage}     & {Parameters}      & {Output}   \\ 
		\hline
		{Conv + ResBlock$\times$2 $+$ Downsampling}      &   {$\left[\begin{array}{l}
				3\times3,128,1 \\
                3 \times 3,128,1 \\
				3 \times 3,128,1 \\
				3 \times 3,128,2     
			\end{array} \right] $}     & {$\frac{H}{2}\times \frac{W}{2}$} & {ResBlock$\times$3 $+$ Upsampling} & {$\left[\begin{array}{l}
				3\times3,512,1\\
				3\times3,512,1 \\
				3\times3,512,1 
			\end{array} \right]$}   & {$\frac{H}{16}\times \frac{W}{16}$ }    \\
		{ResBlock$\times$2 $+$ Downsampling}     &   {$\left[\begin{array}{l}
				3 \times 3,128,1 \\
				3 \times 3,128,1 \\
				3 \times 3,128,2
			\end{array}\right]$}      & {$\frac{H}{4}\times \frac{W}{4}$}    &   { ResBlock$\times$3 $+$ Upsampling} &      {$\left[\begin{array}{l}
				3\times3, 512, 1\\
				3 \times 3,512,1 \\
				3 \times 3,512,1 
			\end{array} \right]$}        & {$\frac{H}{8}\times \frac{W}{8}$}    \\
		{ResBlock$\times$2 $+$ Downsampling}     &   {$\left[\begin{array}{l}
				3 \times 3,128,1 \\
				3 \times 3,128,1 \\
				3 \times 3,256,2
			\end{array}\right]$}    &{ $\frac{H}{8}\times \frac{W}{8}$}    & {ResBlock$\times$3 $+$ Upsampling} & {$\left[\begin{array}{l}
				3\times3, 256, 1\\
				3 \times 3,256,1 \\
				3 \times 3,256,1 
			\end{array} \right] $} & {$\frac{H}{4}\times \frac{W}{4}$}    \\
		{ResBlock$\times$2 $+$ Downsampling}     &    {$\left[\begin{array}{l}
				3 \times 3,256,1 \\
				3 \times 3,256,1 \\
				3 \times 3,256,2
			\end{array} \right]$}    & {$\frac{H}{16}\times \frac{W}{16}$}    &{ResBlock$\times$3 $+$ Upsampling }  &    { $\left[\begin{array}{l}
				3\times3,256,1\\
				3\times3,256,1 \\
				3\times3,256,1 
			\end{array} \right]  $ }        & 
   {$\frac{H}{2}\times \frac{W}{2}$}     \\
		{ResBlock$\times$2 $+$ Downsampling}    &   { $\left[\begin{array}{l}
				3 \times 3,256,1 \\
				3 \times 3,256,1 \\
				3 \times 3,512,2
			\end{array}  \right]$ }    &{ $\frac{H}{32}\times \frac{W}{32}$}    &  {ResBlock$\times$3 $+$ Upsampling }  & {$\left[\begin{array}{l}
				3 \times 3,128,1\\
				3 \times 3,128,1 \\
				3 \times 3,128,1 
			\end{array} \right]  $}   & {$H\times W$}   \\
		{ResBlock$\times$2}    &   { $\left[\begin{array}{l}
				3 \times 3,512,1 \\
				3 \times 3,512,1 \\
			\end{array}  \right]$ }    &{ $\frac{H}{32}\times \frac{W}{32}$}    &  {ResBlock$\times$3} & {$\left[\begin{array}{l}
				3 \times 3,128,1 \\
				3 \times 3,128,1 \\
				3 \times 3,128,1
			\end{array} \right]$}   & {$H\times W$}  \\
		{--}    &   {--}    &{--}    & {Classifier} & {$\left[\begin{array}{l}
				3 \times 3,1,1 \\
			\end{array} \right]$}   & {$H\times W$}  \\
		\toprule
	\end{tabular}
\end{table*}

\section{Experiment}\label{sec:experiment}
\subsection{Experimental Dataset}
In this section, extensive computer experiments are performed on four CD datasets, namely CD Dataset (CDD) \cite{CDDdataset},  LEVIR-CD \cite{LEVIRdataset},  WHU-CD \cite{ji2018fully} and  Global Very-high-resolution Landslide Mapping (GVLM) \cite{LMdataset}.
\begin{itemize}[leftmargin=*]
\item {\bf CDD} is a publicly available large-scale CD dataset containing image pairs across four seasons. Furthermore, the spatial resolution of its images varies from $3$ to $100$ centimeters/pixel, which enables the dataset to accurately characterize objects of various sizes, ranging from cars to large construction structures. In particular, the CDD covers seasonal changes in natural objects, including a single tree and wide forest areas. Finally, we divide the dataset into a training set of $10,000$ images, a test set of $3,000$ images and a validation set of $3000$ images with each image of size  $256\times 256$ pixels;
\item {\bf LEVIR-CD} is a public large-scale building CD dataset containing $637$ pairs of high-resolution remote sensing images of spatial resolution of $0.5$ meter. After cropping the original images into small non-overlapping  patches of size $256\times 256$ from the original images, we obtained $7120$, $1024$ and $2048$ pairs of patches for training, validation and testing, respectively;
\item {\bf WHU-CD} is a public building CD dataset containing high-resolution aerial images of size spatial resolution of $7.5$ centimeters. After cropping the original images into small non-overlapping patches of size $256\times 256$, we randomly divide the resulting data into a training set of $6096$ patches, a test set of $762$ patches and a validation set of $762$ patches, respectively,
\item {\bf GVLM} is a publicly available CD dataset specifically designed for landslide monitoring. It contains $17$ pairs of high-resolution landslide images for landslide monitoring and mitigation algorithm development and evaluation. Each image was split into patches with a size of $256\times 256$ pixels. Finally, we obtained $4557$, $1518$ and $1518$ pairs of patches for training, validation and testing, respectively.
\end{itemize}

\begin{figure*}
	\centering
	\includegraphics[width=0.90\textwidth]{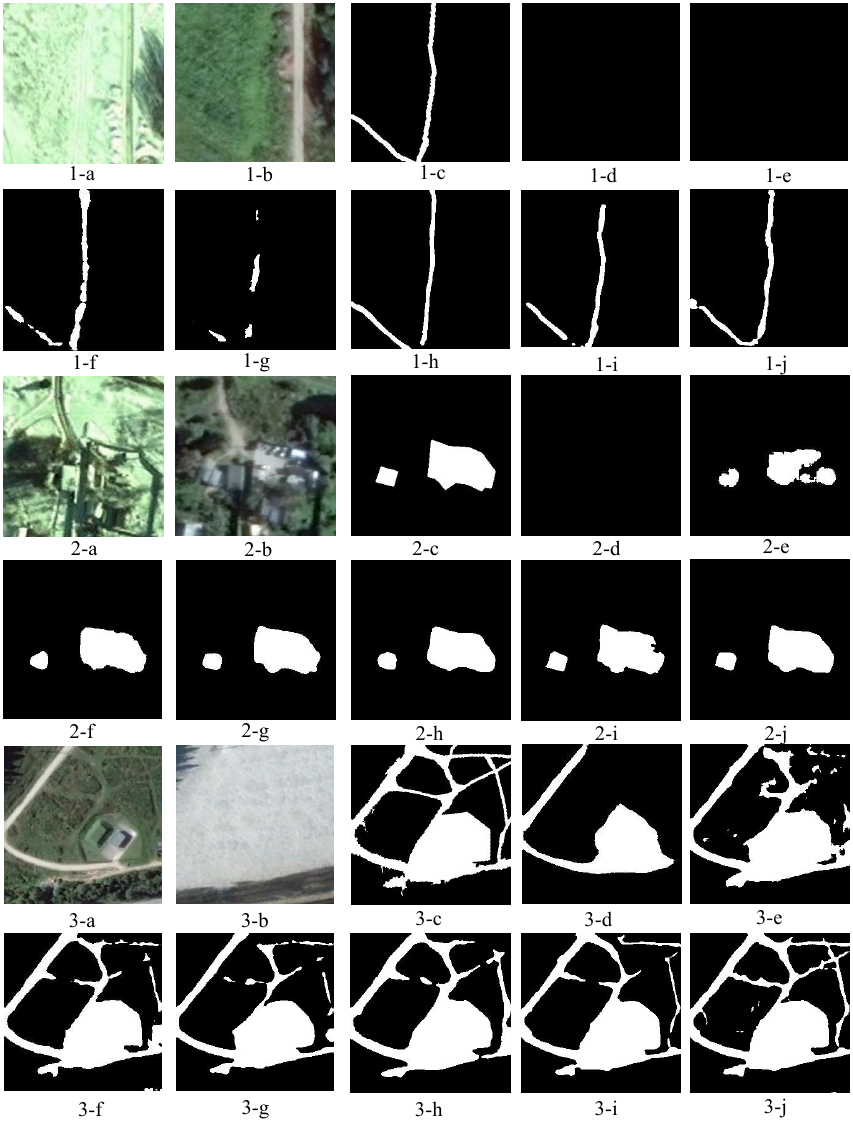}
	\caption{Comparision of different state-of-the-art CD methods on CDD dataset: (a) Pre-change image, (b) Post-change image, (c) Ground-truth, (d) FC-SC, (e) SNUNet, (f) DT-SCN, (g) BIT, (h) AMTNet, (i) DDPM-CD, and (j) the proposed GCD-DDPM.
	}
	\label{fig:CDDFG}
\end{figure*}

\subsection{Implementation Details}
In the proposed model, the noise predictor adopts an encoder-decoder structure based on residual U-Net. The parameters of each convolutional layer in the noise predictor, including kernel size, channel dimension and stride, as well as the output size of each stage, are summarized in Table \ref{table_network}.

Our experiments were implemented using the PyTorch framework empowered by a single NVIDIA GeForce RTX 4090 Ti GPU. 
Furthermore, the stochastic gradient descent (SGD) algorithm with momentum was adopted as the optimization algorithm to train the model with a batch size of $4$. In addition, the momentum and the weight decay parameters are set to $0.99$ and $0.0005$, respectively. Finally, the learning rate was initially set to $0.0001$ before linearly decaying to zero over the course of $200$ epochs. In the inference process, $1000$ diffusion steps were performed.

\subsection{Evaluation Metrics}
The F1-score, Intersection over Union ($IoU$), and overall accuracy ($OA$) are common performance metrics used to evaluate the effectiveness of various models, particularly in tasks such as CD. Each of these metrics provides a different perspective on the performance of the CD model under evaluation. More specifically, the F1-score stands for a balanced assessment of a model's accuracy and completeness. It is a harmonic mean of $Precision$ and $Recall$ and calculated as follows:
\begin{eqnarray}
	F1 = \frac{2}{Recall^{-1} + Precision^{-1}},\\
        Recall = \frac{TP}{TP + FN},\\
	Precision = \frac{TP}{TP + FP},
\end{eqnarray}
where $TP$, $FP$ and $FN$ represent the number of true positives, false positives, and false negatives, respectively.

In addition, IoU is a metric used to evaluate the degree of overlap between two sets, typically the predicted segmentation map and the ground truth. In the context of CD, IoU for the change category is calculated as follows:
\begin{equation}
	IoU = \frac{TP}{TP+FN+FP}.
\end{equation}

$OA$ is a measure of the proportion of correctly classified instances out of the total instances. It is calculated as:
\begin{equation}
	OA = \frac{(TP+TN)}{(TP+TN+FN+FP)}, 
\end{equation}
where $TN$ denotes the number of true negatives.

\subsection{Comparisons}
Comprehensive comparison was performed to compare the proposed approach against five state-of-the-art models reported in the literature, namely three convolution-based techniques, namely FC-SC \cite{cnn2}, SNUNet \cite{skipconnection} and DT-SCN \cite{DT-SCN}, as well as the transformer-based BIT \cite{BIT} and AMTNet \cite{AMTCD}, and the DDPM-based DDPM-CD
\cite{DDPM-CD}. More specifically, FC-SC leverages a Siamese FCN to extract multi-level features while employing feature concatenation as its fusion strategy for bitemporal information \cite{cnn2}. Furthermore, SNUNeT extracts high-resolution high-level features by jointly exploiting the Siamese network and NestedUNet \cite{skipconnection} whereas DT-SCN \cite{DT-SCN} exploits the inter-dependency between channels and spatial positions using a dual-attention module for more comprehensive feature-level fusion. Notably, AMTNet \cite{AMTCD} advances bi-temporal CD in high-resolution optical remote sensing by integrating ConvNets with transformers in a Siamese network. Finally, BIT facilitates the identification of changes of interest while excluding irrelevant alterations by incorporating the transformer architecture into the CD task \cite{BIT} while the DDPM-CD performs CD by employing the DDPM as a feature extractor.

\subsection{Experiment on CDD Dataset }
The CDD dataset, with emphasis on seasonal changes in remote sensing images, was specifically designed for CD algorithm development by incorporating images acquired across different seasons. Fig.~\ref{fig:CDDFG}(a)-(c) show the pre- and post-change images as well as the ground truth. As illustrated in Fig.~\ref{fig:CDDFG}(a)-(c), the CD images include different types of land cover and land use types, such as urban areas, and agricultural fields in different seasons, which enables the evaluation of CD algorithms across different spectral characteristics and scenarios. 

Figs.~\ref{fig:CDDFG}(d)-(i) illustrates the predictive capabilities of all methods under consideration. Inspection of Fig.~\ref{fig:CDDFG} shows that the seasonal changes in vegetation and other land cover types can lead to areas of pseudo-changes in the CD maps, which makes it challenging to identify the actual changes from those caused by seasonal variability. For instance, AMTNet tends to overestimate change areas, potentially leading to noises, as seen in Fig.~\ref{fig:CDDFG}(h). Conversely, DDPM-CD integrates DDPM-based feature extraction into a discriminative framework, offering precise delineation in some cases. However, it achieves results with obvious omissions, indicated by Fig.~\ref{fig:CDDFG}(i). It is evident in Fig.~\ref{fig:CDDFG}(j) that the proposed GCD-DDPM could better preserve the actual boundaries of changed objects while generating predicted change maps of greater detail.

\begin{table}[t]
	\centering
	\caption{The quantitative experimental results ($\%$) on the CDD. THE VALUES IN BOLD ARE THE BEST.}
	\begin{tabular}{cccccc}\hline
		\multirow{1}{1cm}{\textbf{Method}}&
		\textbf{Recall}&\textbf{Precision}&\textbf{OA}&\textbf{F1}&\textbf{IoU}\cr
		\hline
		FC-SC & 71.10 & 78.62 & 94.55 & 74.67 & 58.87\\
		SNUNet    & 80.29 & 84.52 & 95.73 & 82.35 & 69.91 \\
		DT-SCN & 89.54 & 92.76  & 97.95 & 91.12  & 83.59\\
		BIT   & 90.75 & 86.38 & 97.13 & 88.51 & 79.30 \\
        AMTNet & 93.87 & 94.96 & 98.68 & 94.41 & 89.37 \\
		DDPM-CD  & 94.43 & \textbf{95.05} & 98.81 & 94.74 & 90.05\\
		\hline
		Proposed GCD-DDPM & \textbf{95.10} & 94.76 & \textbf{98.87} & \textbf{94.93} & \textbf{90.56} \\
		\hline
	\end{tabular}
	\label{tab:CDDTa}
\end{table}

Table~\ref{tab:CDDTa} shows the performance of the six models under consideration in terms of OA, F1-score, Precision, and IoU. Inspection of Table~\ref{tab:CDDTa} reveals that the proposed GCD-DDPM achieved the best performance in most performance metrics. In particular, the conventional DDPM-CD generated a slightly higher precision, i.e., $95.05\%$, at the cost of worse $FN$ and subsequently recall performance. In contrast, the proposed GCD-DDPM achieved comparable precision performance of $94.76\%$ as well as impressive recall performance.

\begin{table}[t]
	\centering
	\caption{The quantitative experimental results ($\%$) on the WHU-CD. THE VALUES IN BOLD ARE THE BEST.}
	\begin{tabular}{cccccc}
		\hline
		\multirow{1}{1cm}{\textbf{Method}}&  \textbf{Recall}&\textbf{Precision}&\textbf{OA}&\textbf{F1}&\textbf{IoU}\cr
		\hline
		FC-SC     & 86.54  & 72.03 & 98.42 & 78.62  & 64.37 \\
		SNUNet   & 81.33 & 85.66  & 98.68 & 83.44 & 71.39 \\
		DT-SCN & 93.60 & 88.05  & 99.32 & 90.74 & 83.55\\
		BIT  & 87.94 & 89.98 & 99.30 & 88.95 & 81.53 \\
        AMTNet & 91.26 & 92.46 & 99.35 & 91.86 & 84.63 \\
		DDPM-CD   & 92.05 & 92.71  & 99.37 & 92.38 & 85.84 \\
		\hline
		Proposed GCD-DDPM & \textbf{92.29} & \textbf{92.79} & \textbf{99.39} & \textbf{92.54} & \textbf{86.52} \\
		\hline
	\end{tabular}
	\label{tab:WHUTA}
\end{table}

\begin{table}[h]
	\centering
	\caption{The quantitative experimental results ($\%$) on the LEVIR-CD. THE VALUES IN BOLD ARE THE BEST.}
	\begin{tabular}{cccccc}
		\hline
		\multirow{1}{1cm}{\textbf{Method}}&   \textbf{Recall}&\textbf{Precision}&\textbf{OA}&\textbf{F1}&\textbf{IoU}\cr
		\hline
		FC-SC     & 77.29 & 89.04 & 98.25  & 82.75 & 69.95 \\
		SNUNet   & 84.33 & 88.55 & 98.70 & 86.39 & 76.11 \\
		DT-SCN & 87.03 & 85.33 & 98.65 & 86.17 & 75.09\\
		BIT  & 87.85 & 90.26 & 98.83 & 89.04 & 80.12 \\
        AMTNet  & 89.39 & \textbf{91.54} & 99.03 & 90.45 & 82.59 \\
		DDPM-CD    & 89.67 & 91.39 & 99.06 & 90.52 & 82.73\\
		\hline
		Proposed GCD-DDPM & \textbf{91.24} & 90.68 & \textbf{99.14} & \textbf{90.96} & \textbf{83.56}\\
		\hline
	\end{tabular}
	\label{tab:LEVIRtable}
\end{table}

\begin{table}[h]
	\centering
	\caption{The quantitative experimental results ($\%$) on the GVLM. THE VALUES IN BOLD ARE THE BEST.}
	\setlength{\tabcolsep}{2mm}{
		\begin{tabular}{cccccc}
			\hline
			\multirow{1}{1cm}{\textbf{Method}}&
			\textbf{Recall}&\textbf{Precision}&\textbf{OA}&\textbf{F1}&\textbf{IoU}\cr
			\hline
			FC-SC & 89.53 & 74.47 & 96.47 & 81.31 & 67.24\\
			SNUNet  & 91.67 & 87.22 & 98.97 & 89.39 & 80.85\\
			DT-SCN & 92.95 & 82.05 & 97.16 & 87.16 & 77.31\\
			BIT   & 91.15 & 87.39 & 98.70 & 89.23 & 83.76 \\
           AMTNet   & 90.23 & 93.70 & 90.05 & 91.93 & 85.03 \\
			DDPM-CD  & 93.26 & \textbf{93.92} & 99.31 & 93.59 & 87.29 \\
			
			\hline
			Proposed GCD-DDPM & \textbf{94.25} & 93.79 & \textbf{99.32} & \textbf{94.02} & \textbf{89.09} \\
			\hline
	\end{tabular}}
	\label{tab:LMTA}
\end{table}

\subsection{Experiment on WHU-CD Dataset}
\begin{figure*}
	\centering
	\includegraphics[width=0.90\textwidth]{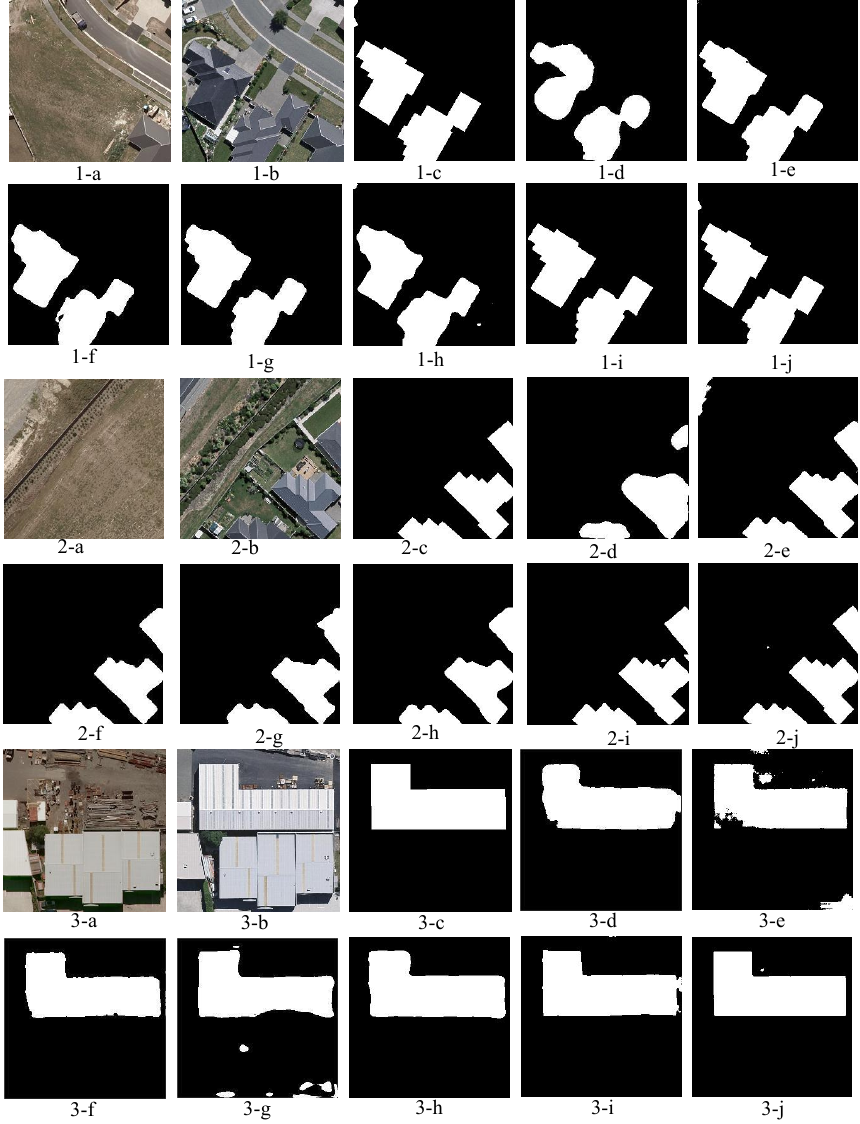}
	\caption{Comparision of different state-of-the-art CD methods on WHU-CD dataset: (a) Pre-change image, (b) Post-change image, (c) Ground-truth, (d) FC-SC, (e) SNUNet, (f) DT-SCN, (g) BIT, (h) AMTNet, (i) DDPM-CD, and (j) the proposed GCD-DDPM.}
	\label{fig:WHUFG}
\end{figure*}

We repeated the experiments above on the WHU building CD dataset. The WHU dataset primarily focuses on urban changes, capturing various types of alterations such as construction, demolition, and land cover transformations. In particular, the WHU dataset contains intricate urban structures and diverse land cover types, which may lead to pseudo-changes and misclassifications in the CD maps. From the visual results in Fig.~\ref{fig:WHUFG} and quantitative metrics in Table~\ref{tab:WHUTA}, it is evident that other methods face challenges in urban change detection. For instance, methods like SNUNet and BIT, despite their higher Recall values, exhibited over-detection of changes with lower Precision values as evidenced in the misclassified areas in Figs.~\ref{fig:WHUFG}(e) and (g). In Fig.~\ref{fig:WHUFG}(i), the DDPM-CD results showed a cleaner detection of changes as compared to SNUNet and BIT, with fewer false positives. However, it struggled to capture the full spectrum of urban changes, leading to occasional omissions of subtle changes. In contrast, the proposed GCD-DDPM achieved impressive performance by effectively mitigating noise and pseudo-changes while preserving the internal compactness of urban objects. Finally, the proposed GCD-DDPM demonstrated superior performance in accurately detecting urban changes in terms of F1 score, OA, and IoU as compared to other existing methods.

\subsection{Evaluation on LEVIR-CD Dataset}
\begin{figure*}
	\centering
	\includegraphics[width=0.90\textwidth]{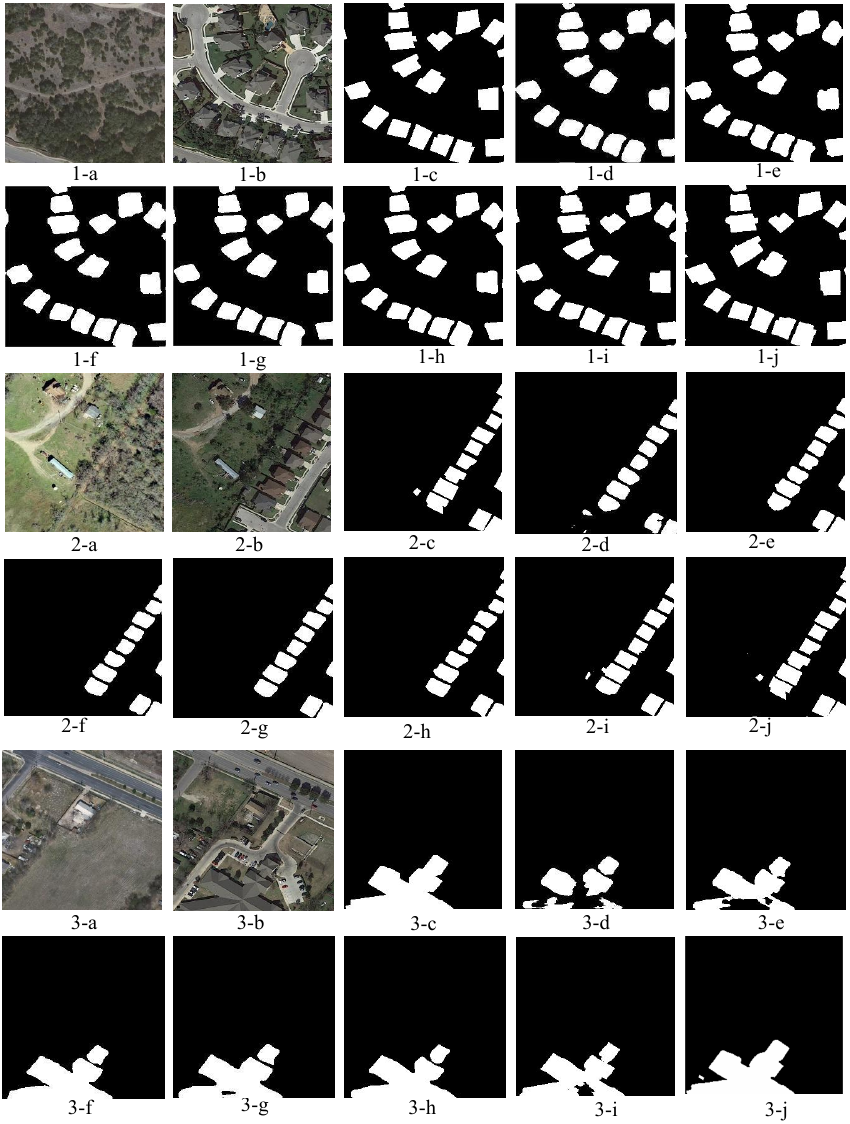}
	\caption{Comparison of different state-of-the-art CD methods on LEVIR-CD dataset: (a) Pre-change image, (b) Post-change image, (c) Ground-truth, (d) FC-SC, (e) SNUNet, (f) DT-SCN, (g) BIT, (h) AMTNet, (i) DDPM-CD, and (j) the proposed GCD-DDPM.}
	\label{fig:LEVIRFG}
\end{figure*}

Next, we performed similar experiments on the LEVIR-CD dataset designed to detect building changes of various scales. As seasonal and illumination changes in the bitemporal images often result in large spectral variability, areas of pseudo-changes commonly occur in the CD maps. This problem can be observed from the small building targets in Fig.~\ref{fig:LEVIRFG} where pseudo-changes were caused in the building roofs while many small building targets were ignored. For example, inspection of Fig.~\ref{fig:LEVIRFG}(h) shows slight overestimation of changed areas with smoothed boundaries for ATMNet. Conversely, Fig.~\ref{fig:LEVIRFG}(i) shows some areas of omissions, leading to incomplete change maps for DDPM-CD. The proposed GCD-DDPM could better preserve the internal compactness of building objects while suppressing noise as compared to other existing models. In particular, it accurately detected most of the small targets while suppressing pseudo-changes.

Table~\ref{tab:LEVIRtable} shows AMTNet with the highest Precision at $91.54\%$. However, GCD-DDPM strikes a good balance between Precision and Recall. This balance is crucial for accurately detecting small structures while minimizing false positives. As a result, GCD-DDPM surpassed AMTNet in overall accuracy and more adeptly addressed spectral variability-induced pseudo-changes in CD tasks.

\subsection{Experiments on the GVLM Dataset}
\begin{figure*}
	\centering
	\includegraphics[width=0.90\textwidth]{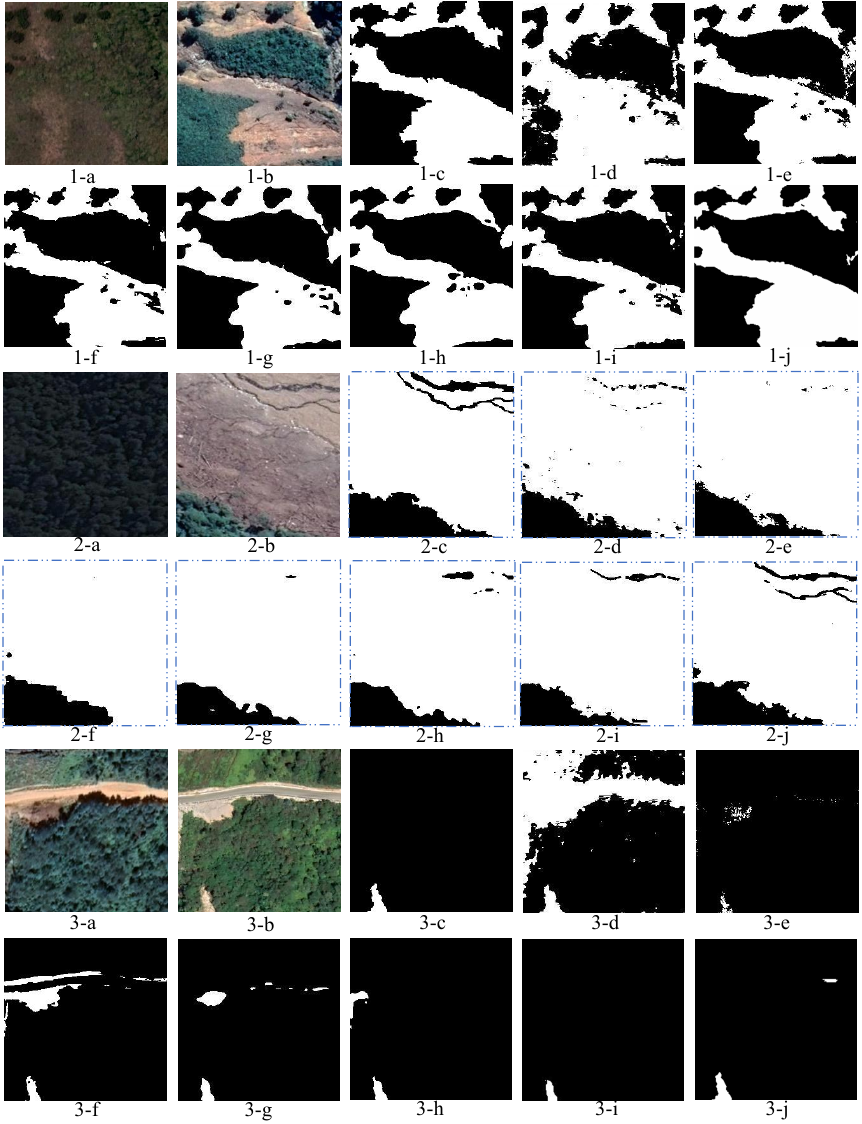}
	\caption{Comparision of different state-of-the-art CD methods on GVLM dataset: (a) Pre-change image, (b) Post-change image, (c) Ground-truth, (d) FC-SC, (e) SNUNet, (f) DT-SCN, (g) BIT, (h) AMTNet, (i) DDPM-CD, and (j) the proposed GCD-DDPM.
	}
	\label{fig:LMFG}
\end{figure*}

Finally, we performed experiments on the GVLM dataset designed for detecting landslide areas from bitemporal images by quantifying land cover changes. The dataset consists of various types of landslides of irregular shapes occurring in regions under different land cover conditions. As illustrated in Fig.~\ref{fig:LMFG}, the CD maps generated by the existing models contained salt-and-pepper noise, especially on the boundaries of landslide objects of irregular shapes. Methods such as SNUNet (Fig.~\ref{fig:LMFG}(f)) and BIT (Fig.~\ref{fig:LMFG}(g)) struggled with the irregular shapes of landslides, leading to noisy CD maps with unclear boundaries. The GCD-DDPM method (Fig.~\ref{fig:LMFG}(j)), in contrast, demonstrated superior performance in detecting the precise edges of landslide changes, offering clean and unambiguous change maps with significantly reduced noise.

The quantitative results in Table~\ref{tab:LMTA} further confirm the superior performance of GCD-DDPM in landslide detection on the GVLM dataset. Table~\ref{tab:LMTA} shows that AMTNet achieved a higher Precision and a diminished Recall due to the inadequate detection of land cover changes. In contrast, GCD-DDPM excelled in recognizing diverse types of changes while alleviating noises, which is evident in its superior IoU and F1 values.

Through the analysis of the experimental results, it can be found that DDPM-CD achieves acceptable results on all datasets compared with other discriminative models because it incorporates a pre-trained DDPM for feature extraction. Nevertheless, its CNN-based structure is still insufficient in discriminative representation learning, leading to some errors of false alarms and missed detections. As a discriminative model, it focuses on explicit classification based on pixel-wise representations and does not fully exploit the iterative refinement and generative capabilities intrinsic to DDPM. In contrast, the superior performance of GCD-DDPM can be attributed to its remarkable generative nature. Specifically, GCD-DDPM adopts an iterative inference process by progressively refining change maps to retain fine-grained details. This adaptive calibration ensures dynamic adaptation to diverse land cover conditions. 
Furthermore, GCD-DDPM integrates change-aware feature representations at multiple levels to guide the noise prediction process and enhance the reliability of change map generation. Therefore, accurate and robust CD maps can be obtained.

\begin{figure}
	\centering
	\includegraphics[width=0.50\textwidth]{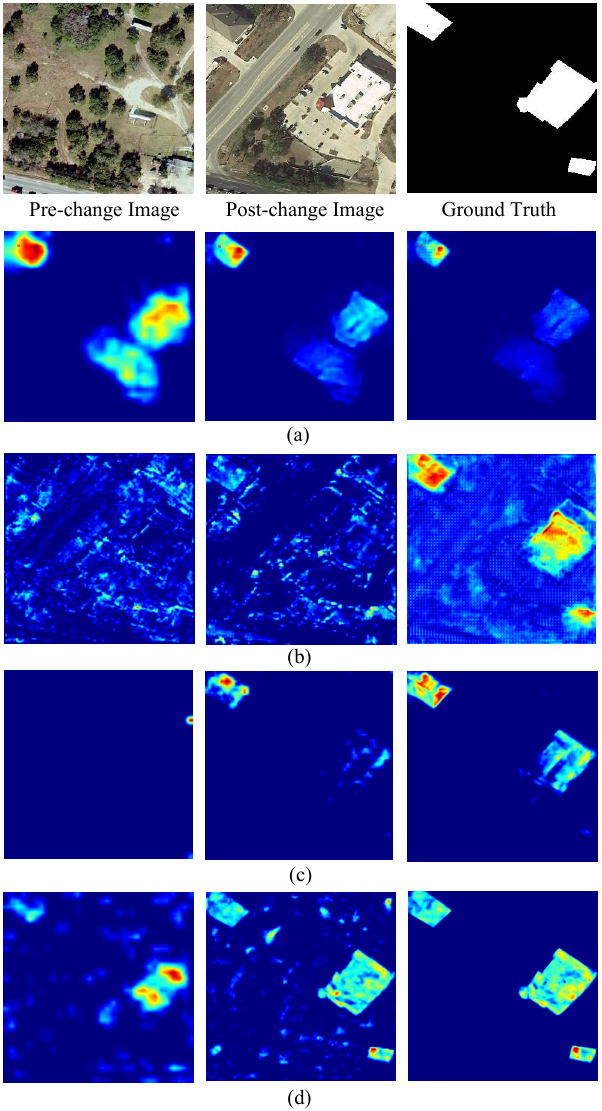}
	\caption{In the heatmaps, (a), (b), (c), and (d) respectively stand for the multi-scale feature maps in the decoder of FC-SC, BIT, DDPM-CD, and GCD-DDPM methods. The three columns of the visualization results represent the heatmaps corresponding to the feature maps at the smallest, intermediate, and largest scales in the decoder, respectively.}
	\label{fig:CAM}
\end{figure}

\subsection{Visualization via Gradient-Based Localization}
To better illuminate the performance enhancements achieved, the Gradient-weighted Class Activation Mapping (Grad-CAM) technique reported in \cite{Grad-CAM} was employed to visually examine the feature map output from each decoder layer. However, unlike image classification tasks that assign a single class label to each image, CD of bitemporal images involves labeling each pixel. To cater to this difference, we adopted a modified Grad-CAM approach proposed in \cite{xianping_CAM} for CD, enabling visualization of classification decisions made by the proposed GCD-DDPM for each pixel. In particular, the modified Grad-CAM technique can discover the essential locations on the feature map for the final decision by tracking the gradient information flow. Consequently, class-discriminative locations in the Grad-CAM maps can show higher scores. 

The images on the first row of Fig.~\ref{fig:CAM} are the pre-change image, post-change image and the ground truth, respectively. We show the modified Grad-CAM heatmaps derived from the localization decoder module using FC-SC, BIT, DDPM-CD and the proposed GCD-DDPM in the next four rows in Fig.~\ref{fig:CAM}. Furthermore, the three columns of the visualization results represent the heatmaps of the smallest, intermediate, and largest scales in the decoder, respectively. In particular, a black dot was marked in the change region. It is important to note that brighter pixels have higher scores and are more likely to be classified as change regions. It is observed in Fig.~\ref{fig:CAM} that the feature maps generated by the existing FC-SC, BIT and DDPM-CD were not sufficiently representative for the recognition of the changed region, particularly for the high-resolution shallow features that retain fine-grained details. In contrast, the proposed GCD-DDPM generated more discriminative and change-aware feature maps at all levels. This is because the proposed GCD-DDPM can enhance the representation and discrimination capabilities of the shallow features while preserving rich local structures. As a result, the proposed GCD-DDPM is capable of handling targeted ground objects of different scales more effectively than its transformer and CNN-based counterparts.

\subsection{Ablation Study}
\begin{table}[t]
	\centering
	\caption{Ablation experiment results($\%$) On the WHU-CD dataset.THE VALUES IN BOLD ARE THE BEST}
	\label{tab:123}
	\begin{tabular}{@{}ccccccc@{}}
		\toprule
	NSSE & SCALE1 & SCALE2 & SCALE3 & OA & F1 & IoU \\ \midrule
		\checkmark & &  & \checkmark & 99.32 & 91.32 & 84.35 \\
		\checkmark & & \checkmark & \checkmark & 99.37 & 92.18 & 85.93 \\ 
		\checkmark & \checkmark & \checkmark & \checkmark & \textbf{99.39} & \textbf{92.54} & \textbf{86.52}\\ 
		\midrule
		 & \checkmark & \checkmark & \checkmark & 99.33&91.52&84.69\\
		\bottomrule
	\end{tabular}
\end{table}

Next, we conducted two sets of ablation experiments on the proposed GCD-DDPM. First, we investigated the importance of multi-scale DCE information for CD performance. We considered difference feature maps of three scales that are referred to as SCALE1, SCALE2 and SCALE3 with SCALE3 being the smallest scale. Two ablation experiments were conducted. In the first ablation experiment, the difference information of the medium scale (SCALE2) and the smallest scale (SCALE3) is employed. In contrast, the second ablation experiment directly feeds the information from the smallest scale (SCALE3) into the last encoding stage without employing the proposed NSSE. 

As shown in the first three rows of Table~\ref{tab:123}, the results of the ablation experiments reveal that the performance of the original model outperformed the other two ablated models, indicating that the multi-scale DCE information played a crucial role in enhancing the CD results. In particular, the large improvement was derived from incorporating both SCALE2 and SCALE3 (the second row), as compared to the case with SCALE3 only (the first row), which confirmed the importance of the medium scale (SCALE2) information.

In the second set of ablation experiments, the proposed NSSE module was removed from the proposed CDDMP. As shown in the last row of Table~\ref{tab:123}, the removal of the proposed NSSE incurred noticeable performance degradation in all three performance metrics as compared to the complete GCD-DDPM. In particular, the F1-score suffered from a large degradation of $1.02\%$ whereas IoU $1.83\%$. This ablation experiment confirmed the crucial role played by the NSSE module in enhancing the model's performance in CD tasks. 

\subsection{Efficiency}

\begin{table}[h]
\centering
\caption{Comparison on computation complexity}
\begin{tabular}{lccc}
 \toprule
 Method & GFLOPs (G) & Para (M) \\
 \midrule
 FC-SC      & 4.73    & 1.35\\
 SNUNet     & 13.79   & 3.01 \\
 DT-SCN      & 13.22   & 31.26\\
 BIT        & 8.75    & 3.04\\
 AMTNet     &  21.56   & 73.63\\
 DDPM-CD    & 838.63  & 434.94  \\
 GCD-DDPM     & 269.52  & 130.83  \\
 \bottomrule
 \end{tabular}
\label{tab:efficiency_metrics}
\end{table}

In order to assess the efficiency of the proposed GCD-DDPM, we evaluated its computational and space complexity against other state-of-the-art CD models, including FC-EF, DT-SCN, BIT, SNUNet, and DDPM-CD in terms of floating-point operations per second (FLOPs) and the number of trainable parameters. As shown in Table~\ref{tab:efficiency_metrics}, the computational complexity of GCD-DDPM is $269.5$ GFLOPs, which is significantly lower than that of DDPM-CD ($838.6$ GFLOPs) but higher than the other non-DDPM based models. In terms of space complexity, GCD-DDPM required $130.8$ million of trainable parameters, which is greater than most of the other non-DDPM-based models. However, it has lower space complexity than DDPM-CD, which utilizes the pretrained DDPM as a feature extractor. Furthermore, as demonstrated in the previous sections, GCD-DDPM's higher computational complexity can be justified by its significantly improved CD performance.

\section{CONCLUSION}\label{sec:conclusion}
In this paper, a denoising diffusion probabilistic model-based CD model called GCD-DDPM has been proposed for remote sensing images. The proposed GCD-DDPM can directly generate high-quality CD maps through end-to-end training, which eliminates the necessity of additional pre-training of diffusion models. In particular, a novel Difference Conditional Encoder (DCE) is designed to guide the CD map generation by exploiting multi-level change information between pre- and post-change images. Additionally, the Variational Inference (VI) used during the DDPM training process enables the model to develop robust feature representations, highly sensitive to subtle differences in data. The GCD-DDPM's inference process further benefits from the adaptive calibration mechanism, which allows for continuous refinement and enhancement of CD accuracy. Furthermore, the Noise Suppression-based Semantic Enhancer (NSSE) employs attention mechanisms to alleviate noises in the change-ware feature representation before they are propagated into the diffusion press. As a result, the proposed GCD-DDPM can dynamically localize and fine-tune the detection results. Extensive experimental results on four CD datasets, namely CDD, WHU-CD, LEVIR-CD and GVLM, have confirmed that the proposed GCD-DDPM outperforms the discriminative CD models.
More specifically, it can significantly reduce pseudo-changes while retaining fine-grained details. Furthermore, the performance degradation of ablated models confirms the effectiveness of multi-level difference features for guiding the CD map generation and the NSSE for alleviating noises. The analysis of gradient-based localization indicates the proposed DCE can improve the discrimination capabilities of feature representations.

The main limitation of GCD-DDPM lies in its substantial memory occupation and time-consuming nature, which can hinder its application in large-scale remote sensing datasets. Moreover, its iterative inference approach may not align with scenarios requiring real-time responses. Therefore, we tend to introduce non-Markovian diffusion processes, which allow for simulation in smaller steps to greatly enhance the sample efficiency of the GCD-DDPM. On this basis, it will be applied to cross-domain remote sensing semantic segmentation tasks, leveraging its generative and discriminative capability.

\small
\bibliographystyle{IEEEtranN}
\bibliography{references}
\end{document}